\documentclass[runningheads]{llncs}
\usepackage[T1]{fontenc}
\usepackage{graphicx}
\usepackage{graphicx}
\usepackage{latexsym}
\usepackage{listings}
\lstnewenvironment{prompt}[2][] 
  {\lstset{
           basicstyle=\ttfamily\footnotesize,
           frame=single,
           breaklines=true,
           breakatwhitespace=true,
           caption={#2},
           #1}}%
  {}   
\usepackage{xcolor}
\usepackage{adjustbox}

\begin{document}
\title{AI Pedagogy: Dialogic Social Learning for Artificial Agents}
\titlerunning{AI Pedagogy}
%
\author{Sabrina Patania \inst{1}, Luca Annese \inst{1},  Cansu Koyuturk \inst{1},  Azzurra Ruggeri \inst{2} and Dimitri Ognibene\inst{1,3}}
\institute{University of Milan-Bicocca, Milan, Italy\and
Technical University of Munich, Munich, Germany\and
Institute for Cognitive Sciences and Technologies (ISTC-CNR), Rome, Italy\\
\email{\{sabrina.patania,luca.annese1,dimitri.ognibene\}@unimb.it}  \\
\email{c.koyutuerk@campus.unimib.it} \\
\email{azzurra.ruggeri@tum.de}}
\authorrunning{S. Patania et al.}
%
\maketitle              
\begin{abstract}
Large Language Models (LLMs) are highly effective at learning from extensive offline datasets but face significant challenges when acquiring complex knowledge dynamically in online scenarios.
Traditional training paradigms, based on supervised or reinforcement learning, reflect a Piagetian view of independent discovery and rely on vast data and sparse feedback, limiting their adaptability.

Inspired by Vygotsky’s sociocultural theory, this study investigates whether structured pedagogical interactions can enhance the efficiency of online learning in LLMs. 
We introduce a novel  training method where a learner LLM engages in structured teaching dialogues with a knowledgeable LLM teacher to learn a synthetic taxonomy. 
The trained learner then applies this knowledge in downstream tasks, specifically tested in the challenging and well-known 20 Questions Game.

These dialogues not only convey new external knowledge but also actively guide the learner in testing and refining its understanding.
Our approach complements internal reasoning methods and prompt engineering: rather than relying on self-generated chains of thought or manually tailored inputs to refine the understanding and response to a single  request, it introduces enriched and reusable task-specific knowledge  through automatically structured pedagogical interactions.
Unlike fine-tuning or few-shot learning, our method introduces novel domain knowledge without altering model weights or requiring explicit task examples.

Our results show that the AI pedagogy strategy combining teacher explanations with learner-driven questions leads to better acquisition and application of knowledge compared to direct access to structured data. This highlights the potential of pedagogically guided interactions to enhance post-training learning and advance the development of more adaptable and human-aligned collaborative AI systems.

\end{abstract}

\section{Introduction}

Over the past five years, Large Language Models (LLMs) have reshaped natural-language processing, demonstrating an ability to answer open-ended questions, draft code, and even reason across  chains of thought \cite{yao2023react,jaech2024openai,guo2025deepseek} and support flexible robot behaviour\cite{saycan2022arxiv,team2025gemini,colombani2024one}.  Such successes, however, rest on a fundamentally offline recipe: pre-training from terabytes of data.  When an LLM is deployed, it must make the best of whatever structure happened to appear in that corpus; it can refine its latent knowledge only indirectly, through prompt engineering or computationally and data expensive fine-tuning.  The result is an awkward mismatch: humans constantly extend, reorganise, and verify their knowledge during conversation, whereas current LLMs have no principled mechanism for acquiring a new, formally structured body of facts on the fly.

A second mismatch between current AI practice and human cognition is social in nature.  Much of human expertise is transmitted interpersonally: parents, teachers, and peers selectively highlight information, manage cognitive load \cite{sweller1988cognitive}, and adjust explanations to a learner’s moment-by-moment understanding.  This Vygotskian view \cite{vygotsky1978mind} contrasts sharply with the dominant “Piagetian”  stance  in machine learning, where agents are typically expected to discover regularities independently—either from direct interactions with the physical world or from static knowledge repositories \cite{carey2015theories,chi2009active}.  If we aim to develop intelligent systems capable of collaborative interaction—systems that acquire updated knowledge, co-write reports, troubleshoot novel hardware, or assist in medical diagnosis—then relying solely on solitary self-discovery may impose a limiting inductive bias.

Educational psychology provides an alternative.  Decades of work on scaffolding, worked examples, and cognitive-load theory show that judicious pedagogical support helps novices build coherent schemas more quickly and robustly.  Translating those insights to AI raises an intriguing question: can an LLM learn complex knowledge more effectively if we place it in a social learning environment, where another agent takes explicit responsibility for teaching?

\begin{figure}[t]
    \centering    \includegraphics[width=0.6\linewidth]{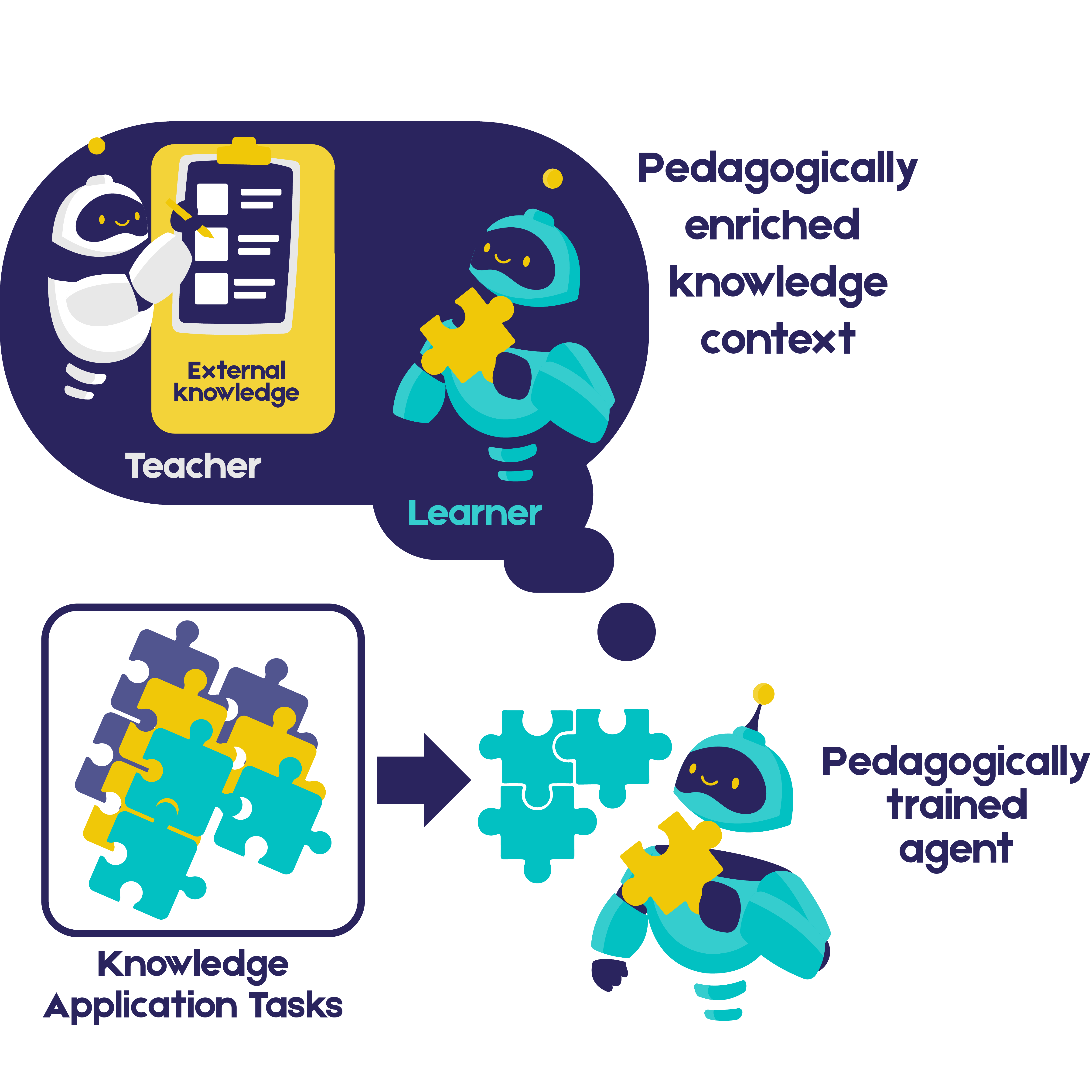}
    \caption{\textbf{Conceptual overview of the experimental setup.} A naïve LLM-based learner acquires an alien taxonomy, external knowledge, through structured pedagogical interactions with a teacher LLM that has direct access to a formal representation of the taxonomy. The pedagogically trained agent is then applied to \textit{Knowledge Application Tasks} and evaluated using a 20 Questions-style deductive task, relying exclusively on knowledge acquired during the pedagogical dialogue, provided as context. This process is flexible and can be iteratively integrated into task execution in human- or AI-in-the-loop scenarios. Notably, it does not require modifying model weights or providing explicit task solution examples.}
    \label{fig:ai_pedagogy}
\end{figure}

In this work, we pursue that question through a controlled study of ontology learning.  We construct fictitious taxonomies of “alien” species, rich enough to require non-trivial structure, yet free from any leakage of real-world priors, and embed them in an interactive task reminiscent of the classic 20-Questions game.  A knowledgeable teacher LLM possesses the ground-truth ontology; a naïve learner must acquire  that ontology solely through dialogue and later apply it in successive tasks (see Fig\ref{fig:ai_pedagogy}). Crucially, we vary the teacher’s pedagogical strategy across four well-studied dimensions in human learning, including top-down exposition, bottom-up induction, learner-driven questioning, and teacher-guided inquiry. 

In addition to isolating these pedagogical styles, we explore mixed strategies resulting in a richer typology of dialogic learning conditions. The experimental setup involves repeated, fixed-length teaching sessions followed by game-based evaluation. Performance is gauged by the learner’s ability to apply the acquired taxonomy in a deductive guessing game. A high-level schematic of the interaction is provided in Figure~\ref{fig:ai_pedagogy}.

Overall, the study indicates that socially interactive, pedagogically steered protocols provide a promising path toward AI agents that can build, verify, and extend structured knowledge in concert with human partners.

\section{Related Work}
Early work in developmental robotics~\cite{lungarella2003developmental} laid the foundation for understanding how robots can acquire cognitive functions, including language~\cite{morse2017there}, through autonomous interaction with their environment, modeling children's developmental processes. These efforts, though primarily Piagetian in approach, highlighted the potential role of language in supporting learning, an idea further developed in epigenetic robotics \cite{lindblom2003social,kozima2000epigenetic,berthouze2003epigenetic}. For instance, \cite{cakmak2010exploiting} examines robot social learning, offering an alternative to costly solitary reinforcement learning \cite{ognibene2019addiction}. Lockerd and Breazeal \cite{LockerdBreazealTutelage2004} introduced an early architecture inspired by human pedagogy at a time when language processing in robots was very limited.
While language’s role in enhancing cognitive capabilities is well studied \cite{petit2012coordinating,sugita2005learning}, its integration within structured pedagogical frameworks for robots remains underexplored. With the emergence of LLMs and related foundation models capable of few-shot learning \cite{brown2020language}, it becomes timely to revisit this.
We investigate how pedagogical interactions between two LLMs, through dialogue, can support knowledge co-construction, in contrast to static, hand tailored prompting approaches, which are often inefficient in robotic contexts especially to encode reusable knowledge \cite{white2023prompt,theophilou2023learning}. Other approaches like fine-tuning, retrieval augmented generation (RAG), or few shot learning have different constraints (e.g. need for expensive weight updates or task solution examples) and can be combined with the proposed AI pedagogy approach.

Insights from developmental and educational psychology offer theoretical grounding for the AI pedagogy framework. Structured, expert-led inquiry (e.g., direct explanation) presents information hierarchically and reduces cognitive load, especially for novices \cite{martin2023integrating,sweller2011cognitive}. In contrast, bottom-up approaches emphasize learner agency and contextual adaptation but risk cognitive overload if unguided \cite{cousins1993enhancing}. Balancing these paradigms is key in designing AI pedagogical systems.

\section{Methods}
Our experiments examine how different pedagogical strategies affect the ability of LLMs to acquire, structure, and apply conceptual knowledge in a grounded interactive setting. The core idea is to model learning as a form of socially mediated knowledge construction, in which a naïve learner agent interacts with a more knowledgeable teacher agent through a pedagogical dialogue. 
\paragraph{Experimental setup.}
We conduct all simulations with GPT-4o accessed through the OpenAI API.\footnote{Temperature is fixed at~0.3 and
\texttt{max\_tokens} at 5000 for every call.}
The procedure consists of three stages: \emph{ontology generation},
\emph{pedagogical interaction}, and
\emph{20-Questions evaluation}.
\paragraph{Ontology generation.}
The core learning materials are entirely invented ontologies of "alien" species, each defined by a small set of categorical features (e.g., diet, habitat, morphology). All ontologies are synthetically generated to ensure no prior grounding in existing data. For each condition, the ontology generated with the help of an LLM remains fixed through both the learning and testing phases.
\paragraph{Pedagogical interactions.}
During training, a knowledgeable teacher LLM and a naïve learner LLM engage in brief, fixed-length dialogues centered on the target alien ontology. 
We include four base strategies, beginning with two primary explanatory framing approaches, monologic Top-down (TD) and Bottom-up (BU), in which the teacher delivers a complete exposition of the ontology while the learner remains passive. In addition, we explore two core dialogic variants, where an initial teacher prompt is followed by a back-and-forth dialogue focused solely on the source of questioning initiative, either from the learner (LQ) or the teacher (TQ) in a unidirectional way.
\begin{itemize}
    \item \textbf{Top-down explanation (TD)}: The teacher leads by introducing high-level feature categories globally (e.g., “There are three kinds of diet...”), then elaborates with grouped species examples to support deductive generalisation.
    
    \item \textbf{Bottom-up induction (BU)}: The teacher begins with specific species descriptions, allowing abstract categories to emerge inductively as the learner is invited to notice patterns across instances.
    
    \item \textbf{Learner-driven questioning (LQ)}: The learner freely asks questions. The teacher responds truthfully and provides one-sentence generic clarification, but does not guide the conceptual framing or structure of the ontology.
    
    \item \textbf{Teacher-guided inquiry (TQ)}: The teacher ask targeted, guiding questions (e.g., “Which locomotion type might suit a crystal desert?”), prompting the learner to form, test, and refine conceptual hypotheses. The teacher provides feedback through confirmation or gentle correction.
\end{itemize}
We also analyze six hybrid scenarios in which the teacher first presents the ontology and then moves through a recurring cycle, explanation, questions, answers, repeated as needed. Two of these scenarios look only at who initiates the questioning (the learner or the teacher) and deliberately omit any explicit explanatory framing. For instance, in \textit{Dialogic Teacher Questions (Dial-TQ)}, the teacher leads the interaction using open-ended prompts, without employing top-down or bottom-up explanation structuring. The other four mixed variants combine explanatory framing (TD or BU) with control of questioning initiative (learner- or teacher-led). For example, in \textit{Dialogic Bottom-Up with Learner Questions (Dial-BU-LQ)}, the teacher introduces knowledge through bottom-up examples while allowing the learner to steer the dialogue through questions.
The \emph{teacher} agent receives the structured ontology as part of its
system prompt.  Thus, for every subsequent turn the teacher has perfect,
verifiable knowledge of every species–feature pair.
 \paragraph{Expert Baseline.}
We define an expert baseline, another GPT-4o instance whose system prompt contains the full ontology
during the 20-Questions game. Apart from that advantage, the baseline plays exactly the same protocol as the learners.
\paragraph{20-Questions evaluation}
After training, the learner’s weights are frozen and the agent is tested in
an automated variant of the classic 20-Questions game.
We generate $50$ candidate sets, each containing eight distinct species
randomly drawn from the ontology.  For each set an \emph{oracle} with full
ontology access secretly selects one target.  The learner asks yes/no
feature questions until it either identifies the target or reaches the
maximum budget of 20 questions.  We record the question count and whether
the final guess is correct.
\section{Results}
An independent samples t-test was conducted to compare the performance of LLM agents trained under different pedagogical strategies against that of the expert baseline agent, which showed an average of 7 questions per trial. Learners in the monologic \textit{TD} strategy (M = 5.15, t(df) = -3.12, p = .0047) and the \textit{Dial-LQ} condition (M = 5.30, t(df) = -3.61, p = .0012) performed significantly better than the expert. In contrast, the LQ, which lacked any expert guidance or framing, performed significantly worse than the expert (M = 12.80, t(df) = 6.95, p < .001). Results for the remaining pedagogical strategies were not statistically significant, although several showed performance trends suggesting potential benefits over the expert baseline.
\begin{figure}[t]
  \centering
  \includegraphics[width=\linewidth]{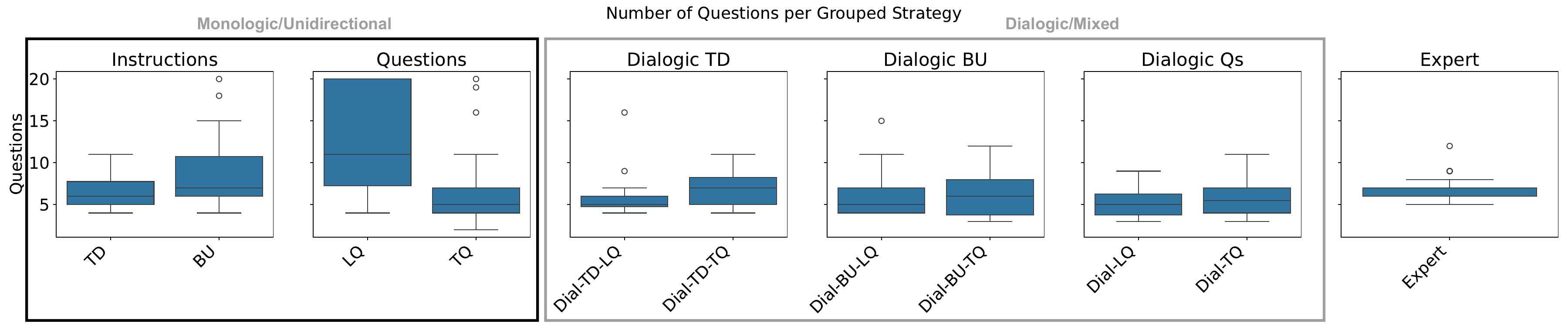}
  \caption{Distribution of questions required to solve the 20-Questions task.  Box-plots show the learner conditions grouped by strategy; the last box on the right is the expert baseline with full ontology access.  Lower values indicate greater efficiency.}
  \label{fig:all}
\end{figure}
\section{Discussion and Conclusions}
Our findings underscore the significant impact of pedagogical strategies on the online acquisition and application of complex, structured, knowledge by LLMs. Notably, the Monologic Teaching Dialogue (TD)  strategy, which supports deductive generalization, and the dialogic TD strategy that interleaves teacher explanations  with learner-initiated questioning, reminiscent of Vygotskian scaffolding, demonstrate superior performance compared to direct access to structured knowledge, even if such format is widely present in training datasets.

The principles underpinning this approach align with foundational ideas in developmental and epigenetic robotics \cite{lungarella2003developmental,morse2017there,LockerdBreazealTutelage2004}. 
Our study builds on these traditions by demonstrating that LLMs, though disembodied, can benefit from structured, socially interactive learning environments. Incorporating rich language interactions and scaffolding techniques may enable the modeling of complex skill acquisition processes akin to those observed in natural social contexts. This progression not only moves us closer to more human-like AI learning systems but also opens avenues for novel contributions from epigenetic robotics to developmental psychology, reinforcing the importance of socially situated, embodied learning~\cite{cangelosi2018babies}.

While inspired by developmental learning theories, our approach is not intended as a direct model of child development. Unlike infants, LLMs already possess an extensive linguistic repertoire. 
The core challenge, and the focus of this work, is enable LLMs to effectively access novel knowledge after training, transforming their passive linguistic capacity into an active tool for incremental knowledge construction.

Traditional methods such as fine-tuning and few-shot learning address this challenge either by modifying model weights or by crafting ad hoc task-specific examples. In contrast, our approach enables the autonomous acquisition and refinement of knowledge through pedagogically structured (simulated) dialogues. Rather than injecting static facts, we support the dynamic co-construction of pedagogically enriched context that is reusable, interpretable, and shaped through interaction. In this sense, our method is orthogonal to prompting and reasoning-based  approaches and complementary to techniques like RAG and in-context learning: where those methods retrieve or inject external content, we construct context dialogically.
%

This dialogue-driven process may offer a more natural and adaptive pathway for real-time knowledge acquisition. It opens the possibility of enhancing LLM capabilities without additional data or model retraining, addressing practical constraints in contemporary AI systems \cite{jaech2024openai,guo2025deepseek}. Moreover, by integrating learner-generated questions to reduce uncertainty during training \cite{patania2024large}, our method helps consolidate new knowledge into structured, reusable internal representations. This may improve the reliability of prompting strategies \cite{white2023prompt,theophilou2023learning}, and strengthen the flexibility of few-shot and zero-shot paradigms \cite{brown2020language}.


A compelling direction for future research lies in understanding the roots of this pedagogical sensitivity in LLMs. Their ability  to respond to socially guided instruction suggests an underexplored propensity to mirror human linguistic behaviors. 
Investigating this capacity could guide the design of more natural pedagogical strategies for real-time knowledge transfer, not only among AI, but also between humans and AIs, including robots, operating in collaborative environments.

%
%
\bibliographystyle{splncs04}
\bibliography{ICSR-AIPedagogy}

\begin{thebibliography}{10}
\providecommand{\url}[1]{\texttt{#1}}
\providecommand{\urlprefix}{URL }
\providecommand{\doi}[1]{https://doi.org/#1}

\bibitem{saycan2022arxiv}
Ahn, M., Brohan, A., Brown, N., Chebotar, Y., Cortes, O., David, B., Finn, C., Fu, C., Gopalakrishnan, K., Hausman, K., Herzog, A., Ho, D., Hsu, J., Ibarz, J., Ichter, B., Irpan, A., Jang, E., Ruano, R.J., Jeffrey, K., Jesmonth, S., Joshi, N., Julian, R., Kalashnikov, D., Kuang, Y., Lee, K.H., Levine, S., Lu, Y., Luu, L., Parada, C., Pastor, P., Quiambao, J., Rao, K., Rettinghouse, J., Reyes, D., Sermanet, P., Sievers, N., Tan, C., Toshev, A., Vanhoucke, V., Xia, F., Xiao, T., Xu, P., Xu, S., Yan, M., Zeng, A.: Do as i can and not as i say: Grounding language in robotic affordances. In: arXiv preprint arXiv:2204.01691 (2022)

\bibitem{berthouze2003epigenetic}
Berthouze, L., Ziemke, T.: Epigenetic robotics—modelling cognitive development in robotic systems (2003)

\bibitem{bertolazzi2023chatgpt}
Bertolazzi, L., Mazzaccara, D., Merlo, F., Bernardi, R.: Chatgpt’s information seeking strategy: Insights from the 20-questions game. In: Proceedings of the 16th International Natural Language Generation Conference. pp. 153--162 (2023)

\bibitem{birjandi2014comparative}
Birjandi, P., Jazebi, S.: A comparative analysis of teachers' scaffolding practices. International Journal of Language and Linguistics  \textbf{2}(3),  154--164 (2014)

\bibitem{brown2020language}
Brown, T., Mann, B., Ryder, N., Subbiah, M., Kaplan, J.D., Dhariwal, P., Neelakantan, A., Shyam, P., Sastry, G., Askell, A., et~al.: Language models are few-shot learners. Advances in neural information processing systems  \textbf{33},  1877--1901 (2020)

\bibitem{cakmak2010exploiting}
Cakmak, M., DePalma, N., Arriaga, R.I., Thomaz, A.L.: Exploiting social partners in robot learning. Autonomous Robots  \textbf{29},  309--329 (2010)

\bibitem{cangelosi2018babies}
Cangelosi, A., Schlesinger, M.: From babies to robots: the contribution of developmental robotics to developmental psychology. Child Development Perspectives  \textbf{12}(3),  183--188 (2018)

\bibitem{carey2015theories}
Carey, S., Zaitchik, D., Bascandziev, I.: Theories of development: In dialog with jean piaget. Developmental Review  \textbf{38},  36--54 (2015)

\bibitem{chen2018learning}
Chen, Y., Chen, B., Duan, X., Lou, J.G., Wang, Y., Zhu, W., Cao, Y.: Learning-to-ask: Knowledge acquisition via 20 questions. In: Proceedings of the 24th ACM SIGKDD International Conference on Knowledge Discovery \& Data Mining. pp. 1216--1225 (2018)

\bibitem{chi2009active}
Chi, M.T.: Active-constructive-interactive: A conceptual framework for differentiating learning activities. Topics in cognitive science  \textbf{1}(1),  73--105 (2009)

\bibitem{colombani2024one}
Colombani, S., Ognibene, D., Boccignone, G.: One to rule them all: natural language to bind communication, perception and action. In: 12th Italian Workshop on Planning and Scheduling (IPS-2024) (2024)

\bibitem{cousins1993enhancing}
Cousins, J.B., Leithwood, K.A.: Enhancing knowledge utilization as a strategy for school improvement. Knowledge  \textbf{14}(3),  305--333 (1993)

\bibitem{degreef2015robots}
De~Greeff, J., Belpaeme, T.: Why robots should be social: Enhancing machine learning through social human-robot interaction. PLoS one  \textbf{10}(9),  e0138061 (2015)

\bibitem{ferdinand2019cognitive}
Ferdinand, V., Kirby, S., Smith, K.: The cognitive roots of regularization in language. Cognition  \textbf{184},  53--68 (2019)

\bibitem{guo2025deepseek}
Guo, D., Yang, D., Zhang, H., Song, J., Zhang, R., Xu, R., Zhu, Q., Ma, S., Wang, P., Bi, X., et~al.: Deepseek-r1: Incentivizing reasoning capability in llms via reinforcement learning. arXiv preprint arXiv:2501.12948  (2025)

\bibitem{ha2023scaling}
Ha, H., Florence, P., Song, S.: Scaling up and distilling down: Language-guided robot skill acquisition. In: Conference on Robot Learning. pp. 3766--3777. PMLR (2023)

\bibitem{jaech2024openai}
Jaech, A., Kalai, A., Lerer, A., Richardson, A., El-Kishky, A., Low, A., Helyar, A., Madry, A., Beutel, A., Carney, A., et~al.: Openai o1 system card. arXiv preprint arXiv:2412.16720  (2024)

\bibitem{kozima2000epigenetic}
Kozima, H., Zlatev, J.: An epigenetic approach to human-robot communication. In: Proceedings 9th IEEE International Workshop on Robot and Human Interactive Communication. IEEE RO-MAN 2000 (Cat. No. 00TH8499). pp. 346--351. IEEE (2000)

\bibitem{lindblom2003social}
Lindblom, J., Ziemke, T.: Social situatedness of natural and artificial intelligence: Vygotsky and beyond. Adaptive Behavior  \textbf{11}(2),  79--96 (2003)

\bibitem{LockerdBreazealTutelage2004}
Lockerd, A., Breazeal, C.: Tutelage and socially guided robot learning. In: 2004 IEEE/RSJ International Conference on Intelligent Robots and Systems (IROS) (IEEE Cat. No.04CH37566). vol.~4, pp. 3475--3480 vol.4 (2004). \doi{10.1109/IROS.2004.1389954}

\bibitem{lungarella2003developmental}
Lungarella, M., Metta, G., Pfeifer, R., Sandini, G.: Developmental robotics: a survey. Connection science  \textbf{15}(4),  151--190 (2003)

\bibitem{martin2023integrating}
Martin, A.J.: Integrating motivation and instruction: Towards a unified approach in educational psychology. Educational Psychology Review  \textbf{35}(2), ~54 (2023)

\bibitem{morse2017there}
Morse, A.F., Cangelosi, A.: Why are there developmental stages in language learning? a developmental robotics model of language development. Cognitive Science  \textbf{41},  32--51 (2017)

\bibitem{nomura2025decentralized}
Nomura, K., Aoki, T., Taniguchi, T., Horii, T.: Decentralized collective world model for emergent communication and coordination. arXiv preprint arXiv:2504.03353  (2025)

\bibitem{ognibene2019addiction}
Ognibene, D., Fiore, V.G., Gu, X.: Addiction beyond pharmacological effects: The role of environment complexity and bounded rationality. Neural Networks  \textbf{116},  269--278 (2019)

\bibitem{ognibene2025scoop}
Ognibene, D., Patania, S., Annese, L., Koyuturk, C., Garzotto, F., Vizzari, G., Ruggeri, A., Colombani, S.: Scoop: A framework for proactive collaboration and social continual learning through natural language interaction andcausal reasoning. arXiv preprint arXiv:2503.10241  (2025)

\bibitem{patania2024large}
Patania, S., Masiero, E., Brini, L., Piskovskyi, V., Ognibene, D., Donabauer, G., Kruschwitz, U., et~al.: Large language models as an active bayesian filter: information acquisition and integration. In: Proceedings of the 28th Workshop on the Semantics and Pragmatics of Dialogue (2024)

\bibitem{petit2012coordinating}
Petit, M., Lall{\'e}e, S., Boucher, J.D., Pointeau, G., Cheminade, P., Ognibene, D., Chinellato, E., Pattacini, U., Gori, I., Martinez-Hernandez, U., et~al.: The coordinating role of language in real-time multimodal learning of cooperative tasks. IEEE Transactions on Autonomous Mental Development  \textbf{5}(1),  3--17 (2012)

\bibitem{van2010scaffolding}
Van~de Pol, J., Volman, M., Beishuizen, J.: Scaffolding in teacher--student interaction: A decade of research. Educational psychology review  \textbf{22},  271--296 (2010)

\bibitem{sugita2005learning}
Sugita, Y., Tani, J.: Learning semantic combinatoriality from the interaction between linguistic and behavioral processes. Adaptive behavior  \textbf{13}(1),  33--52 (2005)

\bibitem{sweller1988cognitive}
Sweller, J.: Cognitive load during problem solving: Effects on learning. Cognitive science  \textbf{12}(2),  257--285 (1988)

\bibitem{sweller2011cognitive}
Sweller, J.: Cognitive load theory. In: Psychology of learning and motivation, vol.~55, pp. 37--76. Elsevier (2011)

\bibitem{taniguchi2020improved}
Taniguchi, A., Hagiwara, Y., Taniguchi, T., Inamura, T.: Improved and scalable online learning of spatial concepts and language models with mapping. Autonomous Robots  \textbf{44}(6),  927--946 (2020)

\bibitem{taniguchi2016symbol}
Taniguchi, T., Nagai, T., Nakamura, T., Iwahashi, N., Ogata, T., Asoh, H.: Symbol emergence in robotics: a survey. Advanced Robotics  \textbf{30}(11-12),  706--728 (2016)

\bibitem{team2025gemini}
Team, G.R., Abeyruwan, S., Ainslie, J., Alayrac, J.B., Arenas, M.G., Armstrong, T., Balakrishna, A., Baruch, R., Bauza, M., Blokzijl, M., et~al.: Gemini robotics: Bringing ai into the physical world. arXiv preprint arXiv:2503.20020  (2025)

\bibitem{team2025kimi}
Team, K., Du, A., Gao, B., Xing, B., Jiang, C., Chen, C., Li, C., Xiao, C., Du, C., Liao, C., et~al.: Kimi k1. 5: Scaling reinforcement learning with llms. arXiv preprint arXiv:2501.12599  (2025)

\bibitem{theophilou2023learning}
Theophilou, E., Koyut{\"u}rk, C., Yavari, M., Bursic, S., Donabauer, G., Telari, A., Testa, A., Boiano, R., Hernandez-Leo, D., Ruskov, M., et~al.: Learning to prompt in the classroom to understand ai limits: a pilot study. In: International conference of the Italian association for artificial intelligence. pp. 481--496. Springer (2023)

\bibitem{thomaz2008teachable}
Thomaz, A.L., Breazeal, C.: Teachable robots: Understanding human teaching behavior to build more effective robot learners. Artificial Intelligence  \textbf{172}(6-7),  716--737 (2008)

\bibitem{thomaz2006experiments}
Thomaz, A.L., Hoffman, G., Breazeal, C.: Experiments in socially guided machine learning: understanding how humans teach. In: Proceedings of the 1st ACM SIGCHI/SIGART conference on Human-robot interaction. pp. 359--360 (2006)

\bibitem{vygotsky1978mind}
Vygotsky, L.S., Cole, M.: Mind in society: Development of higher psychological processes. Harvard university press (1978)

\bibitem{white2023prompt}
White, J., Fu, Q., Hays, S., Sandborn, M., Olea, C., Gilbert, H., Elnashar, A., Spencer-Smith, J., Schmidt, D.C.: A prompt pattern catalog to enhance prompt engineering with chatgpt. arXiv preprint arXiv:2302.11382  (2023)

\bibitem{yao2023react}
Yao, S., Zhao, J., Yu, D., Du, N., Shafran, I., Narasimhan, K., Cao, Y.: React: Synergizing reasoning and acting in language models. In: International Conference on Learning Representations (ICLR) (2023)

\bibitem{yue2025does}
Yue, Y., Chen, Z., Lu, R., Zhao, A., Wang, Z., Song, S., Huang, G.: Does reinforcement learning really incentivize reasoning capacity in llms beyond the base model? arXiv preprint arXiv:2504.13837  (2025)

\end{thebibliography}
\end{document}